\documentclass[10pt,journal,compsoc]{IEEEtran}

\usepackage{graphicx}
\usepackage{amsmath,amssymb,amsfonts}
\usepackage{booktabs,multirow,tabularx,array}
\usepackage{tikz}
\usepackage{xcolor}
\usepackage{algorithm}
\usepackage{algpseudocode}
\usepackage[caption=false,font=footnotesize]{subfig}  
\usepackage{url}
\usepackage[hidelinks]{hyperref}

\title{Toward Reasoning-Centric Time-Series Analysis}

\author{Xinlei~Wang, 
        Mingtian~Tan,
        Jing~Qiu,
        Junhua~Zhao, 
        and Jinjin~Gu\thanks{*Corresponding author: jinjin.gu@insait.ai}
\IEEEcompsocitemizethanks{
\IEEEcompsocthanksitem Xinlei Wang and Jing Qiu are with the University of Sydney, Australia.
\IEEEcompsocthanksitem Mingtian Tan is with the University of Virginia, USA.
\IEEEcompsocthanksitem Junhua Zhao is with The Chinese University of Hong Kong, Shenzhen, and with the Shenzhen Institute of Artificial Intelligence and Robotics for Society, Shenzhen, China.
\IEEEcompsocthanksitem Jinjin Gu is with INSAIT, Sofia, Bulgaria.
}
}

\begin{document}

\IEEEtitleabstractindextext{%
\begin{abstract}
Traditional time series analysis has long relied on pattern recognition, trained on static and well-established benchmarks.  
However, in real-world settings -- where policies shift, human behavior adapts, and unexpected events unfold -- effective analysis must go beyond surface-level trends to uncover the actual forces driving them. 
The recent rise of Large Language Models (LLMs) presents new opportunities for rethinking time series analysis by integrating multimodal inputs. 
However, as the use of LLMs becomes popular, we must remain cautious, asking why we use LLMs and how to exploit them effectively. 
Most existing LLM-based methods still employ their numerical regression ability and ignore their deeper reasoning potential. 
This paper argues for rethinking time series with LLMs as a reasoning task that prioritizes causal structure and explainability.
This shift brings time series analysis closer to human-aligned understanding, enabling transparent and context-aware insights in complex real-world environments.
\end{abstract}

}

\maketitle
\IEEEdisplaynontitleabstractindextext
\IEEEpeerreviewmaketitle

\section{Introduction}
\label{Introduction}
Time series analysis has traditionally been framed as a pattern recognition problem, extracting trends and correlations from observed data. 
Such approaches perform well in numerical fitting but sometimes fail to uncover the causal mechanisms that truly govern temporal evolution.
Real-world time series rarely evolve in isolation. 
%
They are dynamically shaped by external factors, events, policies, and human behaviors. 
Without understanding the underlying logic of temporal dynamics, models cannot adapt to sudden changes driven by these evolving factors. 
As multimodal, context-rich datasets \cite{williams2025context} become increasingly available, reasoning, rather than just pattern extraction, has become both essential and possible.
We argue that time series analysis can be reinterpreted as a reasoning process, in which models infer the underlying logic governing temporal evolution rather than merely extrapolating observed trends.

\textbf{Reasoning over time series involves analyzing observed data in context, leveraging prior domain knowledge to uncover underlying dynamics, explain the emergence of temporal patterns, and ultimately support practical decision-making.}
These go beyond generating the next value or giving a label; they seek to explain why and how changes in context or prior events may shape future dynamics. 
Meanwhile, time series are deeply intertwined with real-world systems, such as energy grids, transportation networks, and healthcare monitoring. 
Thus, improving reasoning in time series is not merely an academic pursuit but a practical necessity for ensuring that data-driven AI translates into reliable and accountable real-world decisions.

Existing studies often fall short in supporting such complex reasoning.
On the one hand, many widely used benchmarks lack detailed contextual information.
For example, datasets like \textit{Australian electricity demand} \cite{Godahewa2021} omit weather and calendar features, while \textit{exchange rate} forecasting \cite{lai2018modeling} often excludes macroeconomic indicators.
Even large-scale datasets such as \textit{M1} \cite{makridakis1982accuracy} consist of over 1000 univariate series without any associated covariates or prediction context.
These settings limit the model’s ability to reason about external drivers and, instead, increase the risk of capturing spurious correlations that lack causal meaning. 
Wang \textit{et al.} \cite{wang2025accuracy} identify an accuracy law revealing that many widely used numerical benchmarks (e.g., ETT \cite{zhou2021informer}, Weather \cite{wu2021autoformer}) for deep time series models have already reached their intrinsic predictability limits. 
Recent efforts \cite{williams2025context} have initiated progress toward context-aware time series benchmarks; however, such multimodal datasets remain rarely utilized in existing research.
On the other hand, even when contextual information is accessible, most existing models still treat forecasting purely as a numerical regression task. 
They lack mechanisms to generate or evaluate intermediate reasoning steps and often fail to maintain contextual coherence. 
This leads to limited interpretability and poor generalization in unseen domains.

As the use of Large Language Models (LLMs) rapidly expands, they offer a new direction for advancing time series analysis. 
Recent studies have employed LLMs either as forecasting backbones \cite{jin2023time,zhou2024one} or by reframing the problems as next-token prediction tasks \cite{gruver2024large,liu2024autotimes}. 
However, their effectiveness in these applications remains under debate. 
Evidence from recent work \cite{tan2024language} indicates that simpler architectures can sometimes match or even surpass LLMs on numerical data, suggesting that many LLM-based forecasting approaches exploit only LLMs’ regression capabilities while neglecting their broader reasoning potential.  
At the same time, studies incorporating textual or contextual information \cite{wang2024news,merrill2024language} demonstrate that LLMs can better identify relevant factors and improve predictive accuracy when provided with meaningful context \cite{williams2025context}. 
Consequently, a key challenge for future research lies in determining \textbf{when} LLMs are genuinely advantageous for time series analysis and \textbf{how} their reasoning can be harnessed most effectively.

Rather than analyzing numerical values in isolation, LLMs excel at processing heterogeneous inputs, such as event descriptions or metadata, to construct causal hypotheses and align forecasts with real-world narratives. 
This positions LLMs not only as predictors but also as cognitive agents. 
To realize this vision, LLMs must be equipped with explicit reasoning capabilities that enable them to connect contextual cues with temporal dynamics. 
A natural question is how such reasoning can be instantiated and trained within current LLM architectures.
Language reasoning capabilities \cite{wei2022chain} provide a foundation for such feasibility.  
Advances in reasoning-aware training strategies -- such as distillation \cite{muennighoff2025s1}, and instruction-aligned frameworks like DeepSeek-R1 \cite{guo2025deepseek} and OpenAI-o1 \cite{jaech2024openai} -- demonstrate that structured, task-oriented reasoning can be taught, distilled, and generalized across domains. 
The ability to model such explicit reasoning is especially important in real-world time series tasks. 
For example, in medicine, reasoning is essential for safe and interpretable decision-making.
Recent work, OpenTSLM \cite{langeropentslm}, illustrates this through ECG question-answering tasks, where the model analyzes electrocardiograms and justifies its diagnosis in natural language. 
Unlike traditional classifiers that output only a label, it produces chain-of-thought (CoT) explanations linking waveform features and patient context to the final decision. 
Such explicit reasoning improves transparency, which is crucial when model predictions directly influence patient care. 
Extending these principles to time series, models capable of reasoning about temporal dependencies will not only enhance accuracy but also yield interpretable, decision-supportive insights. 
This marks a critical step toward the next generation of intelligent time series analysis systems.

\textbf{Therefore, this paper advocates for a greater emphasis on causal relationships and contextual understanding in time series tasks by adopting a reasoning framework with LLMs.}
Such a framework foregrounds causality, interpretability, and multimodal integration to support predictions and decisions in real-world challenges. 
By enabling models to reason about ``why'' and not just ``what,'' it aims to align forecasting systems with real-world dynamics and decision-making needs.

\section{Reasoning for Time Series}
\label{sec-Causality}

Time series data inherently capture temporal cause–and–effect dynamics, where future or latent states depend on historical dependencies and contextual conditions.
\textbf{Time series reasoning} refers to the ability to connect temporal dependencies, detect structural changes, and infer causal mechanisms underlying observed patterns across diverse tasks such as forecasting, classification, and anomaly detection.
Such reasoning allows models to adapt to nonstationary or unseen scenarios.
It handles regime shifts or external shocks rather than merely extrapolating past correlations.

\subsection{Existing Causal Analysis for Time Series}

Although causality is central to reasoning, current causal analyzes in time series fall short of capturing complex real-world dynamics. 
Classical frameworks such as Granger causality~\cite{granger1980testing}, structural VAR models~\cite{sims1980macroeconomics}, Expectation-Maximization~\cite{gong2015discovering}, Nonstationary Causal Temporal Representation Learning~\cite{song2023temporally}, and recent variants such as LOCAL~\cite{zhang2024local} typically operate under a ``\textit{closed system}'' assumption that all relevant variables are fully observed and pre-specified. 
In practice, however, this assumption rarely holds. 
Variable sets are often chosen heuristically without verifying their causal relevance.

Because these methods are confined to ``\textit{closed-set learning}'', they lack perception of external information. 
This limitation poses serious challenges for real-world applications, as many high-order causal mechanisms (e.g., policy changes, social disruptions, or market shocks) are only reflected in multimodal, semantic contexts.
As a result, traditional models often fail to generalize when causal dynamics change abruptly, leading to brittle or misleading forecasts.
Integrating reasoning-oriented LLMs into time series analysis offers a promising direction toward addressing these limitations. 
By jointly interpreting numerical sequences and unstructured contextual data, LLMs can help uncover, reason about, and adapt to evolving causal structures. 

\begin{figure}[t]
\centerline{\includegraphics[width=0.85\columnwidth]{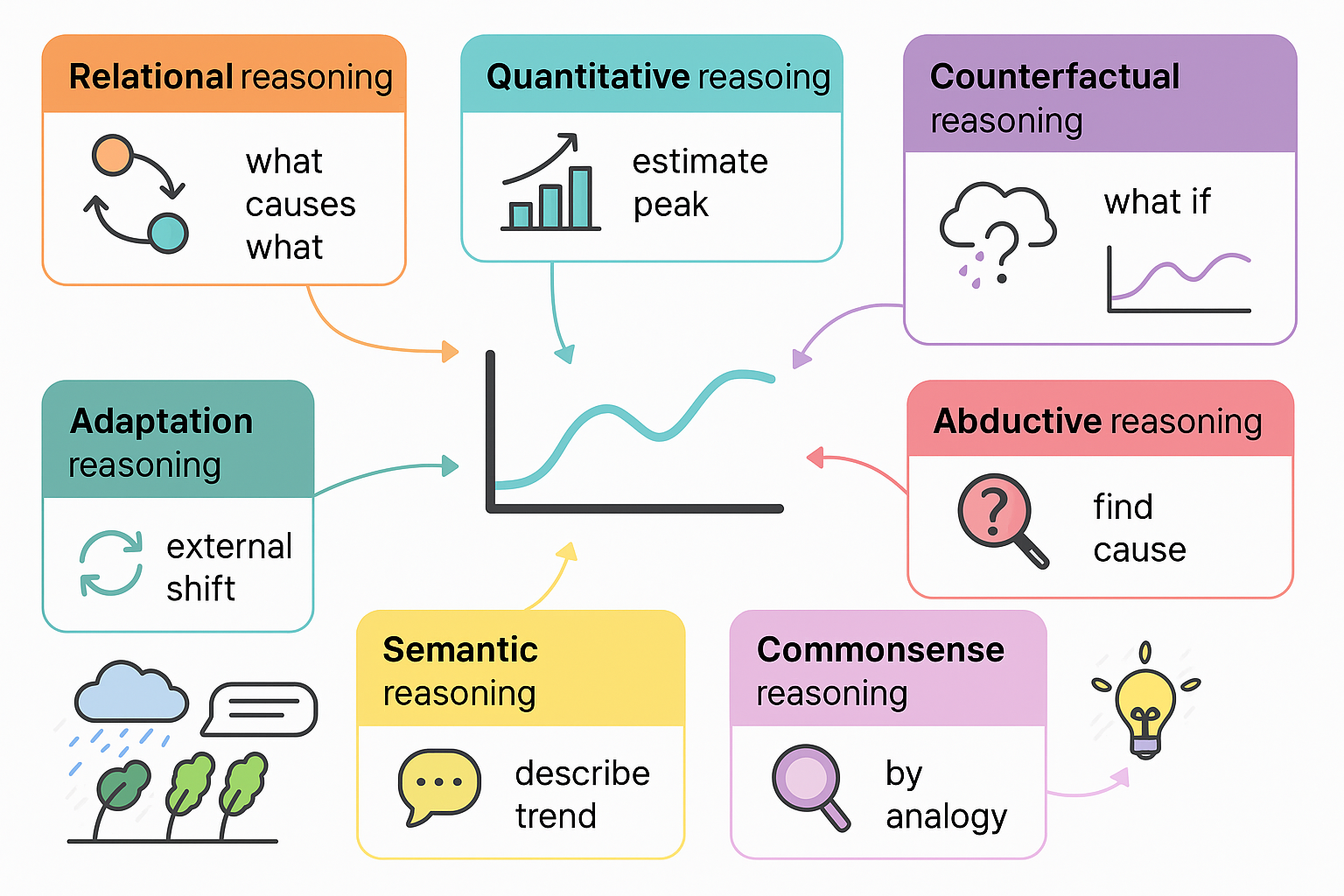}}
\caption{\textbf{Seven types of reasoning essential for Time Series Analysis}.}
\label{nips-reason-type}
\vspace{-4mm}
\end{figure}

\begin{table*}[t]
\centering
\small
\renewcommand{\arraystretch}{1.15}
\caption{A tiered taxonomy of time series reasoning tasks. Levels reflect increasing openness of the environment, heterogeneity of information, and reasoning depth beyond pattern fitting.}
\label{tab:ts_reasoning_tiers}
\begin{tabular}{p{1.5cm}p{2cm}p{3.3cm}p{3.5cm}p{3.5cm}}
\toprule
\textbf{Level} & \textbf{Reasoning Types} & \textbf{Problem Setting} & \textbf{Representative Tasks} & \textbf{Data \& Evaluation} \\
\midrule

\textbf{Level-1 (Structured)} &
Relational, Quantitative &
Closed and stationary systems with predefined variables and clear supervision. &
Point or probabilistic forecasting, classification, and basic anomaly detection. &
Structured numerical series with fixed covariates. Evaluated by standard numerical metrics (e.g., MAE, RMSE, CRPS). \\

\midrule
\textbf{Level-2 (Context-aware)} &
\textit{Level~1 +} Adaptation, Counterfactual, Semantic &
Partially observed or nonstationary systems requiring contextual adaptation. 
&
Regime-shift detection, causal-impact analysis, and time-series captioning. &
Structured and unstructured data (text, policies, or event logs). Evaluated by causal-effect accuracy and robustness under drift. \\

\midrule
\textbf{Level-3 (Open-world)} &
\textit{Level~1 + Level~2 +} Abductive, Commonsense &
Open, multimodal, and nonstationary systems. Models infer latent causes, generalize across domains, and support decision-making. &
Root-cause analysis, policy simulation, strategic planning, retrieval-augmented forecasting, long-form reporting, and temporal question answering. &
Multimodal data (time series + text + images + tables). Evaluated by decision accuracy, cost/benefit metrics, explanation quality, and cross-domain transfer performance. \\

\bottomrule
\end{tabular}
\end{table*}

\subsection{Types of Time Series Reasoning} 
We classify time series reasoning into seven complementary dimensions:

\begin{enumerate}
    \item \textbf{Relational reasoning} uncovers causal and temporal dependencies among observed variables, answering ``what causes what.'' It supports feature attribution and the discovery of structural relationships.

    \item \textbf{Quantitative reasoning} focuses on the magnitude, rate, and timing of change (estimating peaks or inflection points) to support precise predictions and decisions.

    \item \textbf{Counterfactual reasoning} evaluates hypothetical scenarios (``what if'') to estimate the effects of interventions or shocks and to distinguish actual causation.

    \item \textbf{Adaptation reasoning} enables models to adjust under nonstationary or evolving conditions, reasoning about how temporal dynamics should change when external environments shift.

    \item \textbf{Semantic reasoning} translates numerical temporal patterns into natural-language summaries, linking quantitative evidence with contextual meaning.

    \item \textbf{Abductive reasoning} infers plausible but unobserved causes from outcomes, supporting anomaly attribution, diagnostics, and root-cause analysis.

    \item \textbf{Commonsense reasoning} generalizes across domains by analogy or prior experience, allowing models to apply knowledge from historical or similar temporal situations to new ones.
\end{enumerate}

\textbf{Rationale for tiered categorization.} \quad
Time series reasoning spans diverse task complexities and cognitive requirements. To capture this spectrum, we categorize reasoning-oriented time series tasks into three progressive \textbf{levels} (Table~\ref{tab:ts_reasoning_tiers}), reflecting increasing environmental openness, data heterogeneity, and reasoning depth.

\textbf{Level~1 (Structured)} involves closed, well-defined systems where key variables are observable and supervision is sufficient.
Models mainly perform \textit{quantitative} and \textit{relational reasoning}, detecting dependencies, estimating magnitudes, and recognizing temporal structures, which serve as the foundation for higher-level reasoning.

\textbf{Level 2 (Context-aware)} deals with partially observed or evolving systems that require adaptation to external contexts, such as events, textual information, or policy changes. Reasoning extends to \textit{counterfactual} and \textit{semantic} dimensions, emphasizing how interventions and contextual shifts reshape temporal dynamics.

\textbf{Level~3 (Open-world reasoning)} represents the most general setting, where information is multimodal, incomplete, and nonstationary. Models must further extend to \textit{abductive}, \textit{commonsense reasoning} to infer unseen causes, transfer knowledge across domains, and support human-level decision-making. Such tasks include root-cause analysis, policy simulation, long-form reporting, and retrieval-augmented temporal question answering.

Together, these three levels provide a conceptual continuum, clarifying how reasoning complexity scales with task openness and data heterogeneity.

\subsection{Real-World Applications of Time Series Reasoning}
\label{sec:real-world-ts-reasoning}

Real-world applications, which always require Level 3 open-world reasoning ability, must interpret temporal dynamics \emph{in context} and justify predictions for downstream decisions. Below, we summarize representative use cases:

\textbf{Healthcare:} Clinical forecasting requires step-by-step reasoning grounded in patient history and physiological signals.
Models should explain \emph{why} biomarkers (e.g., HRV, glucose) or waveforms (e.g., ECG) change and \emph{how} these relate to disease progression or treatment effects.
This favors explicit, traceable rationales over opaque labels.

\textbf{Energy Systems:} Demand and price forecasting depends on contextual shifts (e.g., weather anomalies, industrial activity, and policy changes).
Reasoning helps distinguish transients from structural regime changes and aligns forecasts with operational decisions (unit commitment, demand response, reserve scheduling).

\textbf{Carbon Markets \& Policy:} Carbon prices and emission trajectories respond to interventions (e.g., carbon tax, Emission Trading Scheme caps).
Reasoning-aware models can assess causal impacts, run counterfactuals, and anticipate portfolio shifts (fuel mix, abatement investment) under policy shocks.

\textbf{Finance \& Economics:} Markets reflect interactions among news, sentiment, and macro indicators.
Integrating text with numerical series enables causal attributions from events to price/volume dynamics and supports adaptive trading or risk management.

\section{LLMs Support Time Series Reasoning}
\label{LLM-TSF}
Given the reasoning demands of complex time series tasks, the next question arises: how to address these diverse needs?
We argue that LLMs are well-suited for this task. 
LLMs have been adapted for reasoning through prompt engineering \cite{marvin2023prompt}, retrieval-augmented generation (RAG) \cite{lewis2020retrieval}, and reasoning-aware training \cite{muennighoff2025s1,guo2025deepseek,jaech2024openai}. 
Recent studies further evaluate their causal reasoning ability \cite{jin2023cladder,jin2023can,jiang2023llm4causal,chen2024causal,chen2025unbiased,liu2025eliciting,zhang2025ac}, employing prompting, fine-tuning, or integration with external causal tools. These works highlight the growing potential of LLMs as general causal reasoners.

\subsection{Existing methods for time series analysis with LLMs}

Recent work has adapted pre-trained LLMs for \textbf{time series forecasting} (TSF) through frozen weights, lightweight tuning, or reprogramming \cite{jin2023time,liu2023unitime,zhou2024one}. 
Models such as TIME-LLM \cite{jin2023time} and TEMPO \cite{cao2023tempo} employ prompt-based temporal representations, while others \cite{zhou2024one,gruver2024large} show that frozen LLMs remain effective when TSF is framed as token prediction.
Further developments include fine-tuned LLMs for time-indexed inputs \cite{ansari2024chronos} and multimodal TSF combining text and temporal signals \cite{jia2024gpt4mts,wang2024chattime}. 
Beyond forecasting, LLMs have been explored for \textbf{classification} and \textbf{anomaly detection}. 
Instruction-tuned models such as \textit{InstructTime} \cite{chen2024instructtime} reformulate classification as prompt–response generation, aligning tokenized sequences with textual labels. 
For anomaly detection, studies investigate zero/few-shot and reasoning-enhanced settings: Zhou \textit{et al.} \cite{zhou2025canllm} evaluate LLMs for anomaly detection, while frameworks such as \textit{AnomalyLLM}, \textit{TriP-LLM}, \textit{SPEAR}, and \textit{CALM} \cite{yu2024anomalyllm,yu2025tripllm,wei2025spear,devireddy2025calm} integrate knowledge distillation, patch-wise tokenization, or semantic evaluation to improve contextual awareness.
However, LLM performance on time series remains debated. Zhou \textit{et al.} \cite{zhou2025canllm} report that CoT prompting offers limited benefits in classification, and Tan \textit{et al.} \cite{tan2024language} show that simpler models often outperform LLMs in purely numerical settings.

Adapting LLMs to time series tasks thus requires explicit reasoning that links causes and effects over time. 
Structured reasoning via Chain-of-Thought or Tree-of-Thought prompting \cite{wei2022chain,yao2023tree} has improved interpretability and transferability in broader domains \cite{muennighoff2025s1,guo2025deepseek}. 
Frameworks such as \textit{OpenTSLM} \cite{langeropentslm} and \textit{GAMETime} \cite{tan2025inferring} demonstrate reasoning-driven approaches that fuse temporal and textual signals for interpretable event inference; however, generalizable reasoning-centric frameworks remain an open challenge.

LLMs have also been deployed as \textbf{reasoning agents} that analyze unstructured contexts and refine predictions through interaction \cite{shinn2023reflexion,cai2023large,cheng2024sociodojo,wang2024news,jiang2025explainable}.
For instance, SocioDojo \cite{cheng2024sociodojo} treats LLMs as collaborative agents for human-like reasoning in financial markets, while News-driven Forecast \cite{wang2024news} and TimeXL \cite{jiang2025explainable} use multi-agent reflection loops to extract causal signals and improve robustness.
Such systems highlight the potential of LLMs to integrate heterogeneous information and perform context-aware time series analysis.

\begin{figure}[t]
\centering
\includegraphics[width=0.45\textwidth]{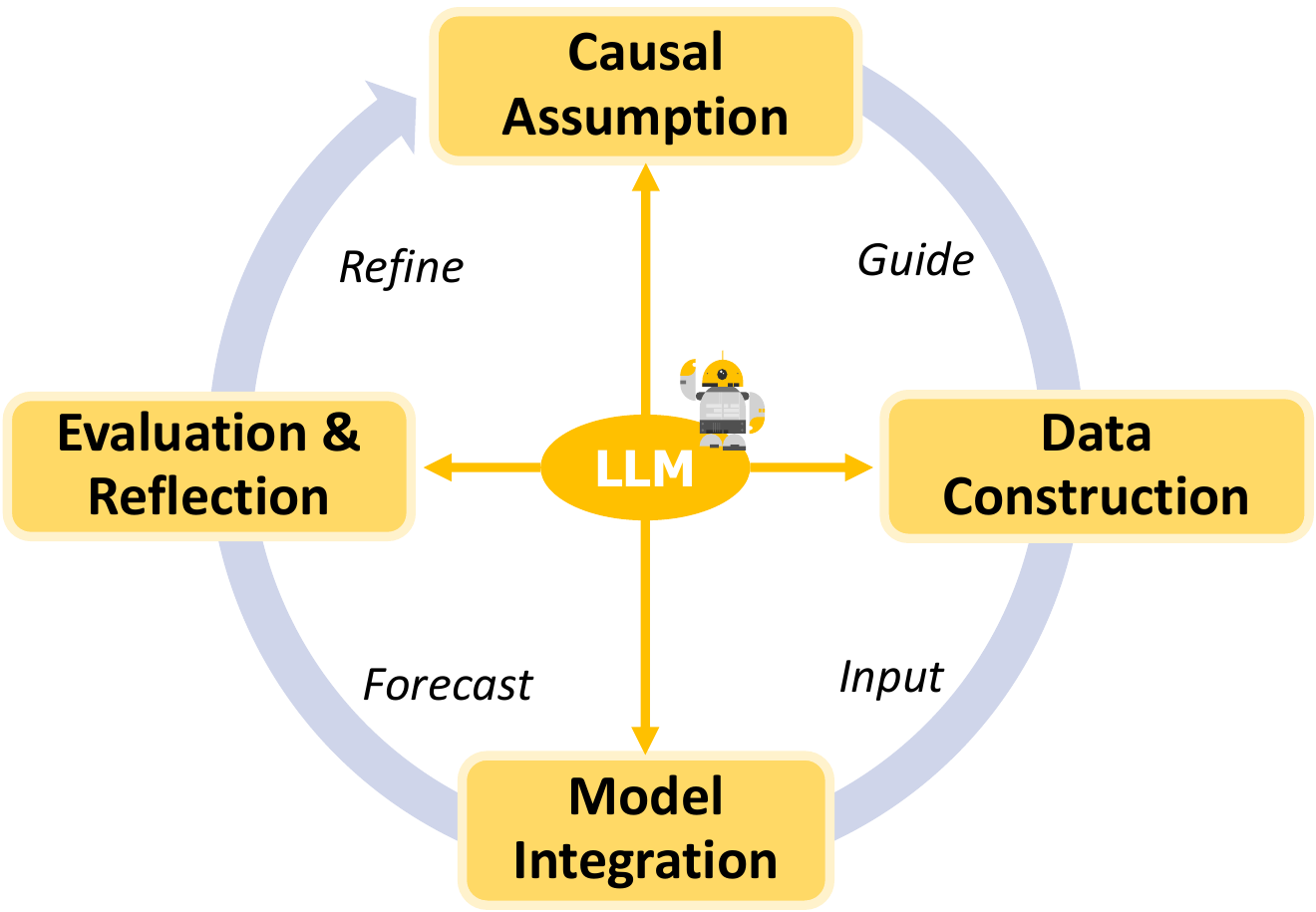}
\caption{An instructive pipeline for time series analysis with LLMs.} 
\label{nips-pipeline}
\end{figure}

\subsection{An instructive pipeline}

We propose an instructive pipeline for \textbf{time series reasoning}, where LLMs act not only as predictors but also as cognitive agents that support causal assumptions, data selection, modeling, and evaluation (Figure~\ref{nips-pipeline}).
Together, these components form a feedback loop that enables models to adapt to evolving causal structures, contextual shifts, and diverse task objectives, such as forecasting, classification, and anomaly detection.

\textbf{Causal assumption.}\quad
Reasoning begins with identifying plausible drivers or explanatory factors underlying temporal behaviors. 
LLMs can autonomously infer causal hypotheses by analyzing domain context, historical patterns, and external knowledge retrieved through RAG frameworks, thereby reducing reliance on manual expertise. 
Prompt-based commonsense reasoning helps elicit potential causal factors (e.g., policies, events, or climate anomalies), while step-by-step CoT prompting supports hypothesis generation and counterfactual reasoning -- exploring what would happen under alternative conditions or interventions.

\textbf{Data construction.}\quad
LLMs dynamically identify, filter, and organize relevant variables based on causal hypotheses and task requirements. 
By parsing metadata, policy documents, or event text, they highlight informative features and suppress noise, enabling models to adapt to evolving contexts and nonstationary dynamics. 
Agent-based systems such as TimeXL \cite{jiang2025explainable} and News-driven Forecast \cite{wang2024news} demonstrate how LLMs can guide feature selection and preprocessing across domains where causal drivers shift over time.

\textbf{Model integration.}\quad
Once causal and contextual insights are established, they can be incorporated into different time-series tasks. 
\begin{enumerate}
    \item \textit{LLM-assisted models:} LLMs convert unstructured context into structured features, explanatory labels, or intervention markers that guide downstream models in forecasting, classification, or anomaly detection. 
    \item \textit{LLM-as-reasoner:} LLMs directly perform task-specific reasoning by embedding historical data, key drivers, and counterfactual cues into prompts, generating both quantitative predictions and textual explanations. Recent work \cite{tan2025inferring} shows that event-aware reasoning enhances both accuracy and interpretability. Supervising outputs with textual rationales -- linking outcomes or detected anomalies to causal chains -- further encourages interpretable reasoning sequences (e.g., ``interest rates rise $\rightarrow$ investment drops $\rightarrow$ demand declines'').
    
    \item \textit{Post-training enhancement:} Beyond task integration, reasoning capabilities can be further strengthened through post-training strategies such as self-distillation \cite{muennighoff2025s1}, reward-aligned fine-tuning \cite{guo2025deepseek}, or reasoning-specific optimization. These methods refine multi-step reasoning and temporal understanding by aligning generated rationales with human-preferred causal logic. They teach models to ``think in steps'' and generalize reasoning skills across diverse temporal tasks.
\end{enumerate}

\textbf{Evaluation and reflection.}\quad
Finally, model outputs are evaluated for semantic plausibility, causal consistency, and explanatory soundness. 
LLMs can verify whether results align with commonsense and assumed drivers, identify contradictory reasoning, and infer new plausible causes when inconsistencies arise (abductive reasoning), as demonstrated in News-driven Forecast \cite{wang2024news}. 
This reflective step closes the reasoning loop, ensuring that predictions, classifications, and anomaly judgments are grounded in transparent, causally coherent explanations.

\begin{figure*}[t]
\centerline{\includegraphics[width=2\columnwidth]{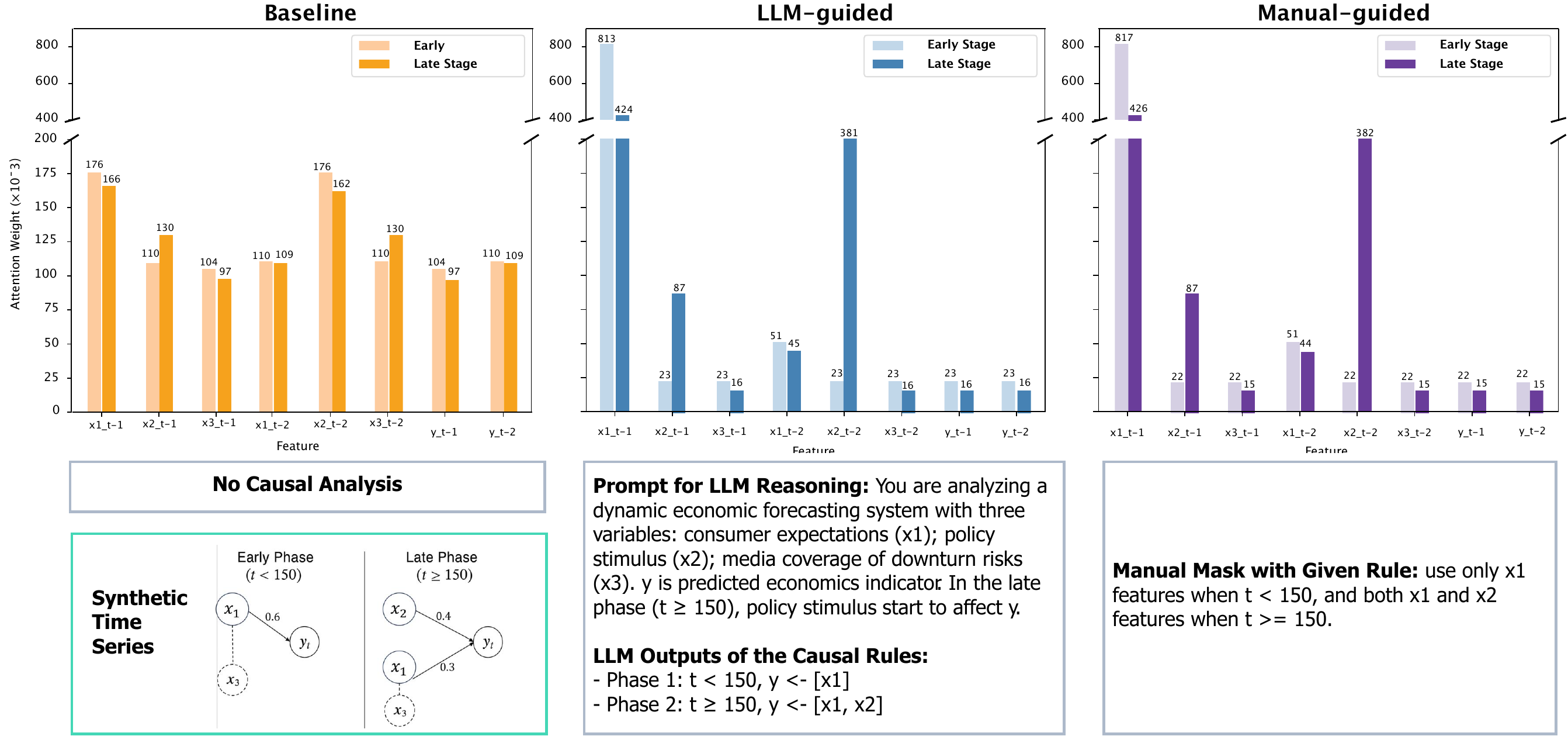}}
\caption{
Toy example illustrating how LLM-guided reasoning improves causal alignment in TSF. We compare attention distributions under three strategies: (1) \textit{Baseline}, with no causal filtering; (2) \textit{LLM-guided}, where LLM interprets textual prompts to generate dynamic attribution masks; and (3) \textit{Manual}, using oracle masks based on known ground-truth structure. 
The \textcolor{gray}{gray boxes} show the input context provided to each strategy: natural language prompt for LLM-guided, hard-coded rule for Manual, and no causal input for Baseline.
The \textcolor{teal}{green box} summarizes the true causal rules used in generating the synthetic time series -- $x_1$ influences $y$ in the early stage ($t{<}150$), and both $x_1$ and $x_2$ influence $y$ in the late stage ($t{\geq}150$).
Results show that only the LLM-guided and Manual models adaptively shift attention from $x_1$ to $x_2$ as causal drivers evolve, while ignoring the spurious $x_3$.
}
\label{nips-toy}
\vspace{-4mm}
\end{figure*}

\subsection{Toy model demonstration}
\label{sec-toy}
To make the idea tangible, we built a small synthetic example in which an LLM interprets textual cues to decide which variables matter over time. The setup imitates a dynamic system whose causal drivers change: early on, one variable dominates; later, a new policy factor takes effect. Rather than hard-coding these shifts, the LLM reads short natural-language prompts (e.g., ``policy influence increases after $t = 150$'') and adapts the model's attention accordingly. As shown in Figure.\ref{nips-toy}, the visualization clearly shows how reasoning guides the system to ignore spurious inputs and refocus on true causes as conditions evolve -- demonstrating, in miniature, how contextual reasoning can reshape time-series forecasting.

\section{Interpretability with Unstructured Data}
\label{Interpretability}

\begin{figure*}[t]
\centerline{\includegraphics[width=2\columnwidth]{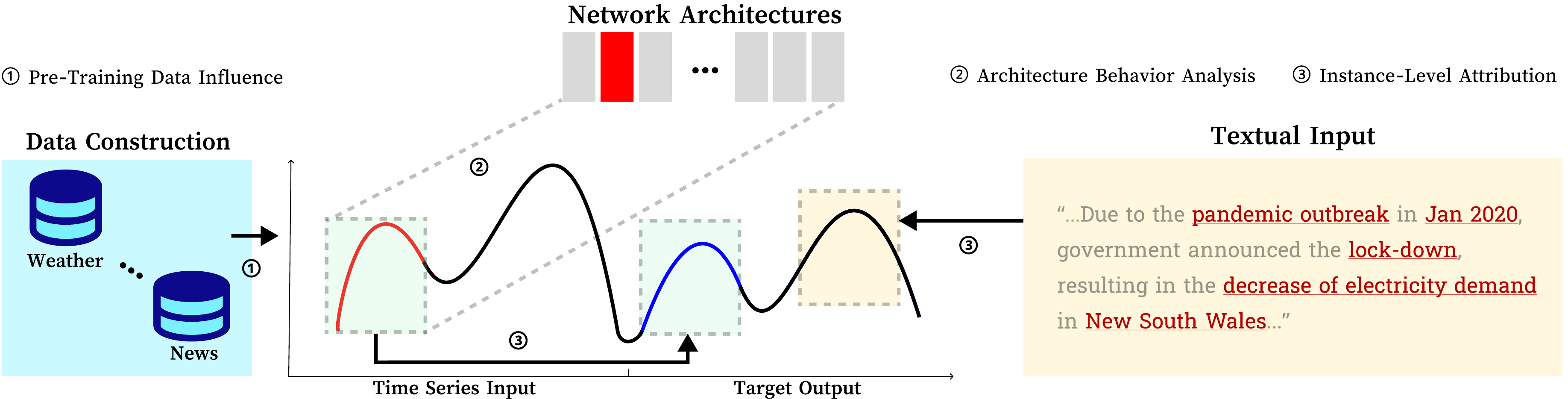}}
\caption{
Interpretability in time series analysis (time series forecasting as an example). This figure illustrates how (1) pre-training data influence, (2) architectural behavior analysis, and (3) instance-level attribution provide insights into the relationship between inputs and their influence on the target output. The red curve represents key features impacting the blue curve, while the red text highlights important textual features relevant to the curve in the yellow-shaded area. The red network layer corresponds to the component responsible for analyzing red curve data (key features).
}
\label{icml-inter}
\end{figure*}

In time series reasoning, interpretability goes beyond post-hoc explanation.
It serves as a tool to assess whether the model's behavior reflects the intended causal reasoning.
It helps answer not just \textit{what} the model predicts, but \textit{why}, which inputs it relies on, how they interact, and whether those relationships align with causal assumptions. 
Despite progress in time series studies, these questions remain unanswered. Current models may exploit frequent patterns rather than capture meaningful causal structures. While recent work has shown that incorporating textual context can improve forecasts, few studies examine whether models actually utilize causal information embedded in that context. Definitions of ``context'' are often vague, and the contributions of specific modalities remain unclear.
As shown in Figure~\ref{icml-inter}, we present a taxonomy of interpretability in TSF that spans instance-level attribution, data influence, and architectural behavior, thereby supporting reasoning validation. 

\textbf{Instance-level analysis.}\quad Instance-level analysis aims to understand the predictions and outputs for a specific instance.
For example, it identifies which input features or time steps play a pivotal role in the forecast and how the model evaluates their relative importance.
Related methods include:
(1) Saliency maps and attribution analysis reveal the contribution of each feature to the final output;
(2) Attention mechanisms assign weights to input components, indicating their relative importance in generating predictions;
and (3) Adversarial examples introduce adversarial perturbations to inputs, drastically changing the prediction results and revealing potential weaknesses in the model's robustness.

\textbf{Data construction and influence.}\quad
The behavior of models largely depends on dataset construction and objective function design, which shape their performance. 
For instance, if a particular category of data is underrepresented in the training set, the model's performance on that category will typically degrade.
This underscores the importance of dataset design in shaping model training and inference in time series analysis. 
Feature engineering plays a crucial role in this process.
Creating and identifying features with rich information can make models more interpretable. 
In addition, it is essential to understand how specific training data points or subsets influence the model's behavior.
This includes identifying impactful data points and assessing sensitivity to noise or outliers. 
Such analyzes improve model transparency, robustness, and interpretability. 
Pinpointing relevant training samples ensures that predictions are trustworthy and reproducible. 

Many data-centric interpretability methods rely on ablation experiments, which involve removing parts of the dataset to observe changes in predictions. 
However, more efficient tools are needed to identify the most relevant features from large feature sets.
``Influence functions'' \cite{koh2017understanding}, for instance, offer another promising avenue: by tracing a model's predictions back to the corresponding training data using robust statistical means, they can identify which training points most strongly influence a particular outcome.
This enhances model interpretability and reliability by offering a deeper understanding of its behavior.

\textbf{Architecture behavior.}\quad Most deep learning research for time series analysis focused on designing improved neural network architectures. 
The reasons behind the success of many neural backbones remain unclear. 
Architectural research often starts with assumptions like ``expanding the temporal receptive field benefits TSF'' and designs methods such as dilated convolutions or deeper layers to test them. 
They then validate the new network's effectiveness through empirical results.
The limited understanding of the internal behavior of deep networks and TSF tasks themselves often makes these assumptions fragile. 
Intuitive as they seem, there is no direct evidence linking ``receptive field expansion'' to better predictions or confirming that architectural changes effectively enhance performance. 
In the absence of a scientifically grounded interpretability perspective, relying solely on final experimental results can lead to biased or incorrect conclusions.

Understanding a prediction architecture's inner workings is key to enhancing interpretability, with potential research avenues including: 
(1) Detailed ablation studies involve systematically removing or substituting key network components to evaluate their impact on performance and interpretability; 
(2) Interpretable model design focuses on developing interpretable architectures, such as modular designs that explicitly separate trend and seasonality; 
and (3) Visualization of learned patterns employs tools like heatmaps or attention maps to represent what the model has learned and how it processes time series data, fostering trust and practical usability.

\section{Outlook}
This paper explores the transformative potential of integrating causal analysis, interpretability, and LLMs into \textbf{time series reasoning}. We advocate for a paradigm shift toward a reasoning-centric framework that leverages LLMs to process multimodal data and infer causal mechanisms underlying temporal dynamics. Such an approach extends beyond forecasting to encompass classification, anomaly detection, and interpretive analysis, enabling models to connect contextual information with temporal behavior. While this paradigm strengthens the ability of time-series models to address dynamic, real-world challenges, it also highlights the need to confront current limitations and open research questions.

\textbf{Strengths.}\quad 
A key advantage of LLMs lies in their capacity to integrate heterogeneous, causally relevant information.  
By combining multimodal signals (textual, visual, and numerical), LLMs enhance interpretability and reasoning consistency across time-series tasks.

(1) \textit{Time series and textual data:}  
Text provides crucial context for temporal reasoning, capturing events, policies, and sentiments that drive system changes.  
Integrating text sources such as news, social media, or reports \cite{wang2024news,li2024cryptotrade,cheng2024sociodojo,liu2024time,merrill2024language,williams2025context} enables LLMs to uncover causal links often invisible to purely numerical models, improving understanding across domains like finance \cite{cecchini2010making}, energy \cite{obst2021adaptive}, and health \cite{zhang2020predicting}.  

(2) \textit{Time series and visual data:}  
Visual modality of time series \cite{shen2025multi} or visual images of other information (such as satellite imagery) provide spatial or environmental context that complements temporal signals.  
Combining visual and temporal inputs \cite{boussif2024improving,gerard2023wildfirespreadts} helps models capture spatiotemporal causality—for example, linking physical conditions to solar irradiance or wildfire dynamics.

\textbf{Challenges.}\quad 
Despite its promise, reasoning-centric time series analysis faces several challenges. First, distinguishing genuine causal signals from confounding correlations -- especially under regime shifts -- remains difficult without structural priors or external validation. Second, multimodal and reasoning-aware inference can be computationally demanding; long prompts and multi-step reasoning traces increase latency. Third, most datasets still lack supervision of the reasoning process, focusing on input–output mappings rather than explicit causal explanations, which hinders the training and evaluation of reasoning quality. Finally, LLMs operating in zero- or few-shot settings may rely on pseudo-causal correlations learned from web data. Promising directions include instruction tuning with domain-grounded causal examples, integration with symbolic causal tools, and exposure to context-specific reasoning traces. Moving forward, we advocate for structured, traceable reasoning as the foundation for generalizable time-series understanding.

\bibliographystyle{IEEEtran}
\bibliography{references}
\bigskip

\end{document}